\pdfoutput=1
\documentclass[10pt,conference,compsocconf]{IEEEtran}

\usepackage{common}
\usepackage[table]{xcolor}
\definecolor{maroon}{cmyk}{0,0.87,0.68,0.32}

\graphicspath{ {figures/} }

\begin{document}

\title{BranchyNet: Fast Inference via Early Exiting from Deep Neural Networks}

% author names and affiliations
% use a multiple column layout for up to three different
% affiliations
 \author{
 \IEEEauthorblockN{Surat Teerapittayanon}
 \IEEEauthorblockA{Harvard University\\
 Email: steerapi@seas.harvard.edu}
 \and
 \IEEEauthorblockN{Bradley McDanel}
 \IEEEauthorblockA{Harvard University\\
 Email: mcdanel@fas.harvard.edu}
 \and
 \IEEEauthorblockN{H.T. Kung}
 \IEEEauthorblockA{Harvard University\\
 Email: kung@harvard.edu}}

% make the title area
\maketitle

\begin{abstract}
Deep neural networks are state of the art methods for many learning tasks due to their ability to extract increasingly better features at each network layer. However, the improved performance of additional layers in a deep network comes at the cost of added latency and energy usage in feedforward inference. As networks continue to get deeper and larger, these costs become more prohibitive for real-time and energy-sensitive applications. To address this issue, we present BranchyNet, a novel deep network architecture that is augmented with additional side branch classifiers. The architecture allows prediction results for a large portion of test samples to exit the network early via these branches when samples can already be inferred with high confidence. BranchyNet exploits the observation that features learned at an early layer of a network may often be sufficient for the classification of many data points. For more difficult samples, which are expected less frequently, BranchyNet will use further or all network layers to provide the best likelihood of correct prediction. We study the BranchyNet architecture using several well-known networks (LeNet, AlexNet, ResNet) and datasets (MNIST, CIFAR10) and show that it can both improve accuracy and significantly reduce the inference time of the network.
\end{abstract}

\IEEEpeerreviewmaketitle

\section{Introduction}
One of the reasons for the success of deep networks is their ability to learn higher level feature representations at successive nonlinear layers. In recent years, advances in both hardware and learning techniques have emerged to train even deeper networks, which have improved classification performance further~\cite{glorot2010understanding,he2015delving}. The ImageNet challenge exemplifies the trend to deeper networks, as the state of the art methods have advanced from 8 layers (AlexNet), to 19 layers (VGGNet), and to 152 layers (ResNet) in the span of four years~\cite{he2015deep,krizhevsky2012imagenet,simonyan2014very}.
However, the progression towards deeper networks has dramatically increased the latency and energy required for feedforward inference. For example, experiments that compare VGGNet to AlexNet on a Titan X GPU have shown a factor of 20x increase in runtime and power consumption for a reduction in error rate of around 4\% (from 11\% to 7\%)~\cite{kim2015compression}. The trade off between resource usage efficiency and prediction accuracy is even more noticeable for ResNet, the current state of the art method for the ImageNet Challenge, which has an order of magnitude more layers than VGGNet. This rapid increase in runtime and power for gains in accuracy may make deeper networks less tractable in many real world scenarios, such as real-time control of radio resources for next-generation mobile networking, where latency and energy are important factors.

\begin{figure}[ht!]
  \centering
    \includegraphics[width=.65\linewidth]{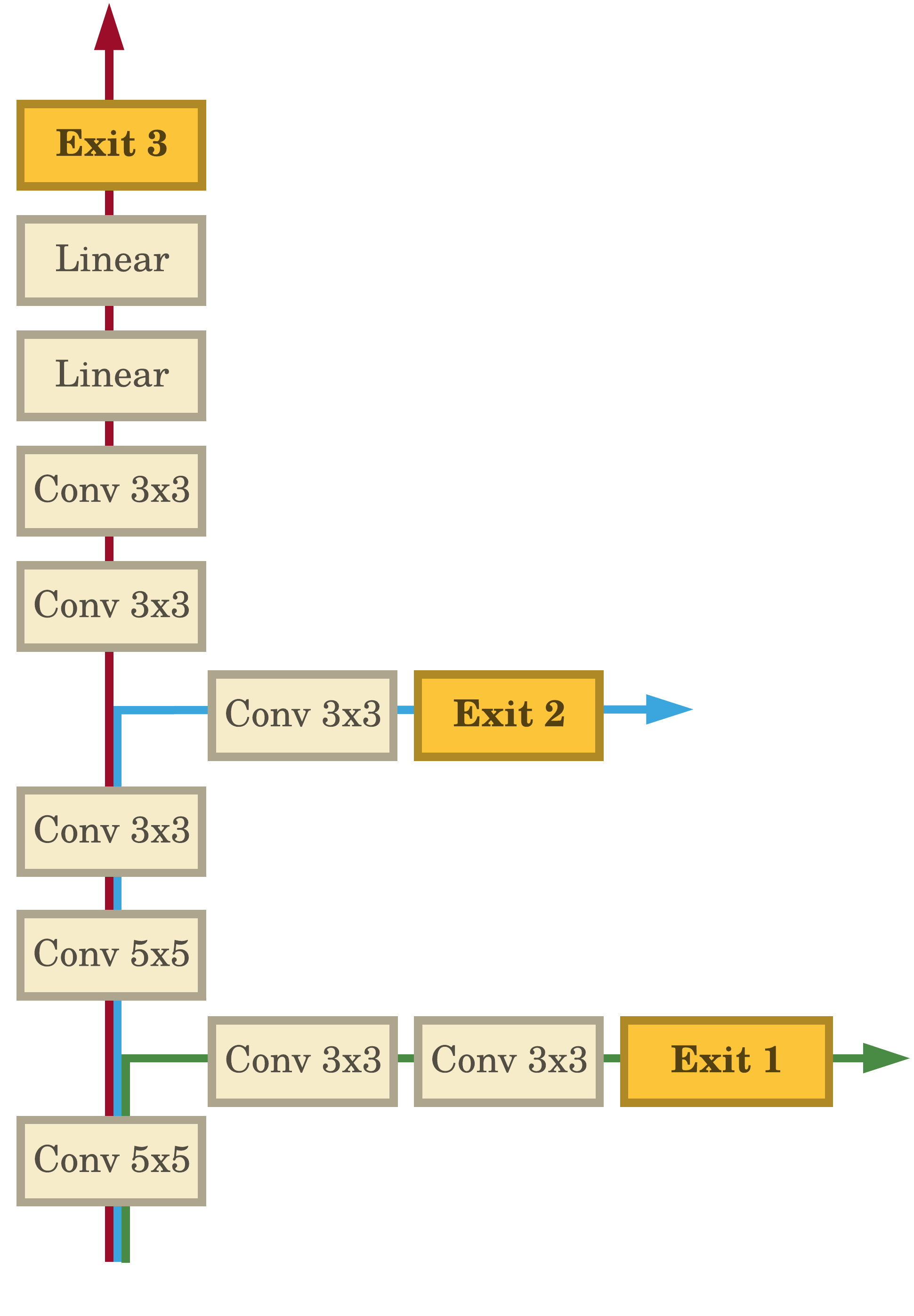}
  \caption{A simple BranchyNet with two branches added to the baseline (original) AlexNet. The first branch has two convolutional layers and the second branch has 1 convolutional layer. The ``Exit'' boxes denote the various exit points of BranchyNet. This figure shows the general structure of BranchyNet, where each branch consists of one or more layers followed by an exit point. In practice, we generally find that it is not necessary to add multiple convolutional layers at a branch in order to achieve good performance.}
  \label{fig:simple_alex}
\end{figure}

To lessen these increasing costs, we present BranchyNet, a neural network architecture where side branches are added to the main branch, the original baseline neural network, to allow certain test samples to exit early. This novel architecture exploits the observation that it is often the case that features learned at earlier stages of a deep network can correctly infer a large subset of the data population. By exiting these samples with prediction at earlier stages and thus avoiding layer-by-layer processing for all layers, BranchyNet significantly reduces the runtime and energy use of inference for the majority of samples. Figure~\ref{fig:simple_alex} shows how BranchyNet modifies a standard AlexNet by adding two branches with their respective exit points.

BranchyNet is trained by solving a joint optimization problem on the weighted sum of the loss functions associated with the exit points. Once the network is trained, BranchyNet utilizes the exit points to allow the samples to exit early, thus reducing the cost of inference. At each exit point, BranchyNet uses the entropy of a classification result (e.g., by softmax) as a measure of confidence in the prediction. If the entropy of a test sample is below a learned threshold value, meaning that the classifier is confident in the prediction, the sample exits the network with the prediction result at this exit point, and is not processed by the higher network layers. If the entropy value is above the threshold, then the classifier at this exit point is deemed not confident, and the sample continues to the next exit point in the network. If the sample reaches the last exit point, which is the last layer of the baseline neural network, it always performs classification.

Three main contributions of this paper are:
\begin{itemize}
\item \textbf{Fast Inference with Early Exit Branches:} BranchyNet exits the majority of the samples at earlier exit points, thus reducing layer-by-layer weight computation and I/O costs, resulting in runtime and energy savings. 
\item \textbf{Regularization via Joint Optimization:} BranchyNet jointly optimizes the weighted loss of all exit points. Each exit point provides regularization on the others, thus preventing overfitting and improving test accuracy. 
\item \textbf{Mitigation of Vanishing Gradients:} Early exit points provide additional and more immediate gradient signal in back propagation, resulting in more discriminative features in lower layers, thus improving accuracy.
\end{itemize}

% In Section~\ref{sec:background} we provide background and related literature. In Section~\ref{sec:formulation}, the BranchyNet structure and training formulation are described in detail. In section~\ref{sec:results}, we evaluate BranchyNet on several well studied/well-known networks (LeNet, AlexNet, ResNet) and datasets (MNIST, CIFAR10) and show that the BranchyNet modification can not only significantly reduce the average runtime of inference  but also improve overall accuracy of the baseline network (about 1\%). BranchyNet provides a CPU and GPU speedup of 5.4/4.7x (LeNet), 1.5/2.4x (AlexNet), and 1.9/1.9x (ResNet). These reductions in runtime over the baseline networks lead to a proportionate savings in energy efficiency. In Section~\ref{sec:discussion}, we provide some analysis in explaining these performance gains of BranchyNet.

\section{Background and Related Prior Work}
\label{sec:background}

LeNet-5~\cite{lecun1998gradient} introduced the standard convolutional neural networks (CNN) structure which is composed of stacked convolutional layers, optionally followed by contrast normalization and maxpooling, and then finally followed by one or more fully-connected layers. This structure has performed well in several image tasks such as image classification. AlexNet~\cite{krizhevsky2012imagenet}, VGG~\cite{simonyan2014very}, ResNet~\cite{he2015deep} and others have expanded on this structure with their own innovative approaches to make the network deeper and larger for improved classification accuracy. %However, deeper and larger models incur additional computations which may not be suitable for fast and energy-efficient inference.

Due to the computational costs of deep networks, improving the efficiency of feedforward inference has been heavily studied. Two such approaches are network compression and implementation optimization. Network compression schemes aim to reduce the the total number of model parameters of a deep network and thus reduce the amount of computation required to perform inference. Bucilua et al. (2006) proposed a method of compressing a deep network into a smaller network that achieves a slightly reduced level of accuracy by retraining a smaller network on synthetic data generated from a deep network~\cite{buciluǎ2006model}. More recently, Han et al. (2015) have proposed a pruning approach that removes network connections with small contributions~\cite{han2015deep}. However, while pruning approaches can significantly reduce the number of model parameters in each layer, converting that reduction into a significant speedup is difficult using standard GPU implementations due to the lack of high degrees of exploitable regularity and computation intensity in the resulting sparse connection structure~\cite{han2015learning}. Kim et al. (2015) use a Tucker decomposition (a tensor extension of SVD) to extract shared information between convolutional layers and perform rank selection~\cite{kim2015compression}. This approach reduces the number of network parameters, making the network more compact, at the cost of a small amount of accuracy loss. These network compression methods are orthogonal to the BranchyNet approach taken in this paper, and could potentially be used in conjunction to improve inference efficiency further.

% \subsection{Implementation Optimization}
% \label{sec:implementation_optimization}
Implementation optimization approaches reduce the runtime of inference by making the computation algorithmically faster. Vanhoucke et al. (2011) explored code optimizations to speed up the execution of convolutional neural networks (CNNs) on CPUs~\cite{vanhoucke2011improving}. Mathieu et al. (2013) showed that convolution using FFT can be used to speed up training and inference for CNNs~\cite{mathieu2013fast}. Recently, Lavin et al. (2015) have introduced faster algorithms specifically for 3x3 convolutional filters (which are used extensively in VGGNet and ResNet)~\cite{lavin2015fast}. In contrast, BranchyNet makes modifications to the network structure to improve inference efficiency.

Deeper and larger models are complex and tend to overfit the data. Dropout~\cite{srivastava2014dropout}, L1 and L2 regularization and many other techniques have been used to regularize the network and prevent overfitting. Additionally, Szegedy et al. (2015) introduced the concept of adding softmax branches in the middle layers of their inception module within deep networks as a way to regularize the main network~\cite{szegedy2015going}. While also providing similar regularization functionalities, BranchyNet has a new goal of allowing early exits for test samples which can already be classified with high confidence.

One main challenge with (very) deep neural networks is the vanishing gradient problem. Several papers have introduced ideas to mitigate this issue including normalized network initialization~\cite{glorot2010understanding,lecun2012efficient} and intermediate normalization layers~\cite{ioffe2015batch}. Recently, new approaches such as Highway Networks~\cite{srivastava2015highway}, ResNet~\cite{he2015deep}, and Deep Networks with Stochastic Depth~\cite{huang2016deep} have been studied. The main idea is to add skip (shortcut) connections in between layers. This skip connection is an identity function which helps propagate the gradients in the backpropagation step of neural network training. 
%Although performance gains of BranchyNet such as being able to improve classification accuracy of the baseline network can in part be attributed to shortcuts, BranchyNet has the distinct design goal of allowing early exits as noted earlier.

Panda et al. \cite{panda2016conditional} propose Conditional Deep Learning (CDL) by iteratively adding linear classifiers to each convolutional layer, starting with the first layer, and monitoring the output to decide whether a sample can be exited early. BranchyNet allows for more general branch network structures with additional layers at each exit point while CDL only uses a cascade of linear classifiers, one for each convolutional layer. In addition, CDL does not jointly train the classifier with the original network. We observed in our paper that jointly training the branch with the original network significantly improve the performance of the overall architecture when compared to CDL.

\section{BranchyNet}
\label{sec:formulation}

BranchyNet modifies the standard deep network structure by adding exit branches (also called side branches or simply branches for brevity), at certain locations throughout the network. These early exit branches allow samples which can be accurately classified in early stages of the network to exit at that stage. In training the classifiers at these exit branches, we also consider network regularization and mitigation of vanishing gradients in backprogation. For the former, branches will provide regularization on the main branch (baseline network), and vice versa. For the latter, a relatively shallower branch at a lower layer will provide more immediate gradient signal in backpropagation, resulting in discriminative features in lower layers of the main branch, thus improving its accuracy.

In designing the BranchyNet architecture, we address a number of considerations, including (1) locations of branch points, (2) structure of a branch (weight layers, fully-connected layers, etc.) as well as its size and depth, (3) classifier at the exit point of a branch, (4) exit criteria for a branch and the associated test cost against the criteria, and (5) training of classifiers at exit points of all branches.  In general, this ``branch'' notion can be recursively applied, that is, a branch may have branches, resulting in a tree structure. For simplicity, in this paper we focus a basic scenario where there are only one-level branches which do not have nested branches, meaning there are no tree branches.

In this paper, we describe BranchyNet with classification tasks in mind; however, the architecture is general and can also be used for other tasks such as image segmentation and object detection.

\subsection{Architecture}
\label{sec:architecture}
A BranchyNet network consists of an entry point and one or more exit points. A branch is a subset of the network containing contiguous layers, which do not overlap other branches, followed by an exit point. The main branch can be considered the baseline (original) network before side branches are added. Starting from the lowest branch moving to highest branch, we number each branch and its associated exit point with increasing integers starting at one. For example, the shortest path from the entry point to any exit is exit 1, as illustrated in Figure~\ref{fig:simple_alex}.

% Given an arbitrary network structure, the BranchyNet architecture can be applied to the network by adding exit branches at various points. Figure~\ref{??} shows and example of BranchyNet architecture. An exit branch consists of multiple layers, such as convolutional and fully-connected layers. We evaluate the choice of the number of layers in a branch and the number of branches in Section~\cite{sec:branches}. 

\subsection{Training BranchyNet}
\label{sec:training}
For a classification task, the softmax cross entropy loss function is commonly used as the optimization objective. Here we describe how BranchyNet uses this loss function. Let $\boldy$ be a one-hot ground-truth label vector, $\boldx$ be an input sample and $\mathcal C$ be the set of all possible labels. The objective function can be written as
\begin{align*}
L(\hat{\boldy}, \boldy;\theta) = 
&-\frac{1}{|\mathcal C|} \displaystyle \sum_{c\in \mathcal C} y_c \log \hat{y_c},
\end{align*}
where 
\begin{align*}
\hat{\boldy}
&= \text{softmax}(\boldz) 
= \frac{\exp(\boldz)}{\displaystyle \sum_{c\in \mathcal C} \exp(z_c)},
\end{align*}
and
\begin{align*}
\boldz =
& f_{\text{exit}_n}(\boldx;\theta),
\end{align*}
where $f_{\text{exit}_n}$ is the output of the $n$-th exit branch and $\theta$ represents the parameters of the layers from an entry point to the exit point.

The design goal of each exit branch is to minimize this loss function. To train the entire BranchyNet, we form a joint optimization problem as a weighted sum of the loss functions of each exit branch
\begin{align*}
    L_{\text{branchynet}}(\hat{\boldy}, \boldy;\theta) = \displaystyle \sum_{n=1}^N w_n L(\hat{\boldy}_{\text{exit}_n}, \boldy;\theta),
\end{align*}
where $N$ is the total number of exit points. Section~\ref{sec:hyper} discusses how one might choose weights $w_n$.

The algorithm consists of two steps: the feedforward pass and the backward pass. In the feedforward pass, the training data set is passed through the network, including both main and side branches, the output from the neural network at all exit points is recorded, and the error of the network is calculated. In backward propagation, the error is passed back through the network and the weights are updated using gradient descent. For gradient descent, we use Adam algorithm~\cite{kingma2014adam}, though other variants of Stochastic Gradient Descent (SGD) can also be used.

% Entropy is calculated at each exit point.

% $\boldz = f_{\text{network}}(\boldx_i,\theta)$

% $\hat{\boldy} = \text{softmax}(\boldz) = \frac{\exp(\boldz)}{\displaystyle \sum_{c} \exp(z_c)}$

% $\text{entropy}(\hat{\boldy}) = \displaystyle -\sum_{j=1}^c {\hat{y}_j \log \hat{y}_j}$

% Backward Propagation

% A training example $\boldx, \boldy$

% Network Parameters $\theta$

% $\boldz = f_{\text{network}}(\boldx_i,\theta)$

% $\hat{\boldy} = \text{softmax}(\boldz) = \frac{\exp(\boldz)}{\displaystyle \sum_{c} \exp(z_c)}$

% $
% L_\text{cross\_entropy}(\hat{\boldy}, \boldy;\theta) = -\frac{1}{c} \displaystyle \sum_{j=1}^c \left[y_j \log \hat{y_j} + (1-y_j ) \log (1-\hat{y_j}) \right]
% $

% $
%  L(\hat{\boldy}, \boldy;\theta) = L_\text{cross\_entropy}(\hat{\boldy}_\text{exit1}, \boldy;\theta) + L_\text{cross\_entropy}(\hat{\boldy}_\text{exit2}, \boldy;\theta) + L_\text{cross\_entropy}(\hat{\boldy}_\text{exit3}, \boldy;\theta)
% $

\subsection{Fast Inference with BranchyNet}
\label{sec:testing}
Once trained, BranchyNet can be used for fast inference by classifying samples at earlier stages in the network based on the algorithm in Figure~\ref{alg:branchy}. If the classifier at an exit point of a branch has high confidence about correctly labeling a test sample $\boldx$, the sample is exited and returns a predicted label early with no further computation performed by the higher branches in the network. We use entropy as a measure of how confident the classifier at an exit point is about the sample. Entropy is defined as
\begin{align*}
    \text{entropy}(\boldy) = \displaystyle \sum_{c\in \mathcal C} y_c \log y_c,
\end{align*}
where $\boldy$ is a vector containing computed probabilities for all possible class labels and $\mathcal C$ is a set of all possible labels.

\begin{figure}[!h]
\begin{algorithmic}[1]
\Procedure{BranchyNetFastInference}{$\boldx,\boldT$}
\For{$n=1..N$}
\State $\boldz = f_{\text{exit}_n}(\boldx)$
\State $\hat\boldy = \text{softmax}(\boldz)$
\State $e\gets \text{entropy}(\hat\boldy)$
% \State $e\gets \text{entropy}(\text{softmax}(\boldz))$
\If{$ e < T_n $}
\State \textbf{return} $\displaystyle \argmax \hat\boldy$
\EndIf
\EndFor
\State \textbf{return} $\displaystyle \argmax \hat\boldy$
\EndProcedure
\end{algorithmic}
\caption{BranchyNet Fast Inference Algorithm. $\boldx$ is an input sample, $\boldT$ is a vector where the $n$-th entry $T_n$ is the threshold for determining whether to exit a sample at the $n$-th exit point, and $N$ is the number of exit points of the network.}
\label{alg:branchy}
\end{figure}

To perform fast inference on a given BranchyNet network, we follow the procedure as described in Figure~\ref{alg:branchy}. The procedure requires $\boldT$, a vector where the $n$-th entry is the threshold used to determine if the input $\boldx$ should exit at the $n$-th exit point. In section~\ref{sec:thresholds}, we discuss how these thresholds may be set. The procedure begins with the lowest exit point and iterates to the highest and final exit point of the network. For each exit point, the input sample is fed through the corresponding branch. The procedure then calculates the softmax and entropy of the output and checks if the entropy is below the exit point threshold $T_n$. If the entropy is less than $T_n$, the class label with the maximum score (probability) is returned. Otherwise, the sample continues to the next exit point. If the sample reaches the last exit point, the label with the maximum score is always returned.

%\subsection{Inference Latency Improvements}
%Since the computation at the later layers and the input of the weights for these layers can be skipped, exiting test samples early provide faster inference, hence improving the speed of inference.

%\subsection{Accuracy Improvements}
%The joint optimization can be view as a regularization. One exit point provides a regularization on another exit point, thus preventing overfitting and improve test accuracy.
%In addtion, early exit points provide additional gradient signal in back propagation, resulting in more discriminative features in lower layers, thus improving the accuracy.

\section{Results}
\label{sec:results}

In this section, we demonstrate the effectiveness of BranchyNet by adapting three widely studied convolutional neural networks on the image classification task: LeNet, AlexNet, and ResNet. We evaluate Branchy-LeNet (B-LeNet) on the MNIST dataset and both Branchy-AlexNet (B-AlexNet) and Branchy-ResNet (B-ResNet) on the CIFAR10 data set. We present evaluation results for both CPU and GPU. We use a 3.0GHz CPU with 20MB L3 Cache and NVIDIA GeForce GTX TITAN X (Maxwell) 12GB GPU.

For simplicity, we only describe convolutional and fully-connected layers of each network. Generally, these networks may also contain max pooling, non-linear activation functions (e.g., a rectified linear unit and sigmoid), normalization (e.g., local response normalization, batch normalization), and dropout. 

For LeNet-5~\cite{lecun1998gradient} which consists of 3 convolutional layers and 2 fully-connected layers, we add a branch consisting of 1 convolutional layer and 1 fully-connected layer after the first convolutional layer of the main network. For AlexNet~\cite{krizhevsky2012imagenet} which consists of 5 convolutional layers and 3 fully-connected layers, we add 2 branches. One branch consisting of 2 convolutional layers and 1 fully-connected layer is added after the 1st convolutional layer of the main network, and another branch consisting of 1 convolutional layer and 1 fully-connected layer is added after the 2nd convolutional layer of the main network. For ResNet-110~\cite{he2015deep} which consists of 109 convolutional layers and 1 fully-connected layer, we add 2 branches. One branch consisting of 3 convolutional layers and 1 fully-connected layer is added after the 2nd convolutional layer of the main network, and the second branch consisting of 2 convolutional layers and 1 fully-connected layer is added after the 37th convolutional layer of the main network. We initialize B-LeNet, B-AlexNet and B-ResNet with weights trained from LeNet, AlexNet and ResNet respectively. We found the initializing each BranchyNet network with the weights trained from the baseline network improved the classification accuracy of the network by several percent over random initialization. To train these networks, we use Adam algorithm with a step size ($\alpha$) of 0.001 and exponential decay rates for first and second moment estimates ($\beta_1,\beta_2$) of 0.99 and 0.999 respectively.

\begin{figure*}[htp]
\centering
\begin{minipage}[t]{.72\linewidth}
\centering
\includegraphics[width=.31\linewidth]{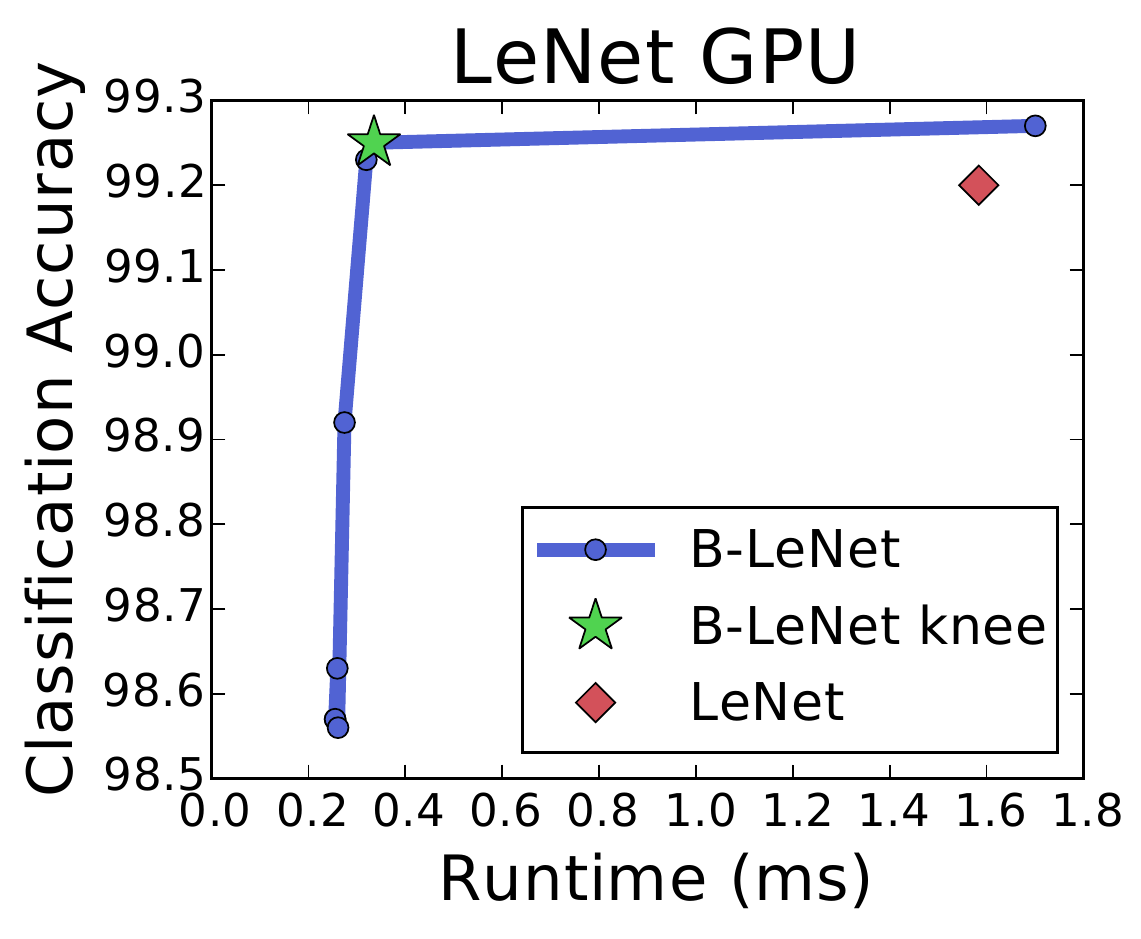}\quad
\includegraphics[width=.31\linewidth]{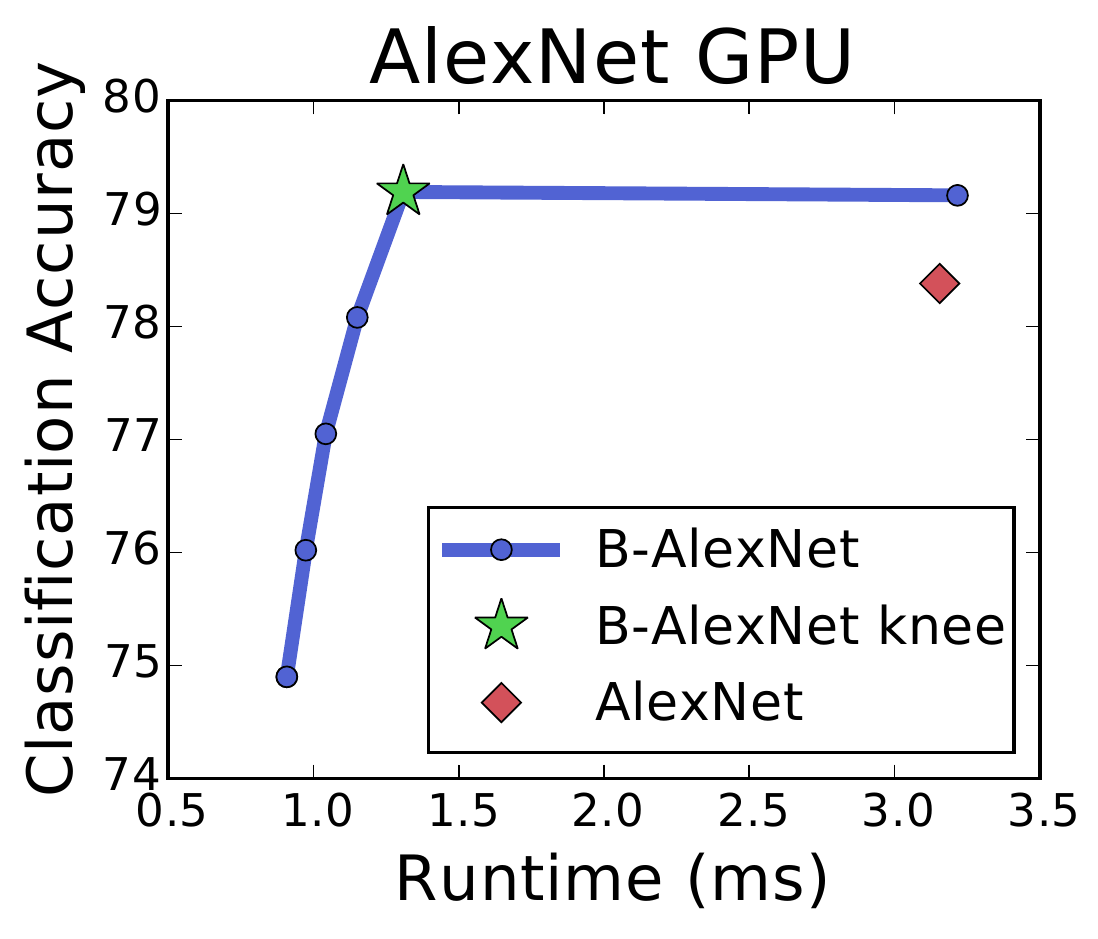}\quad
\includegraphics[width=.31\linewidth]{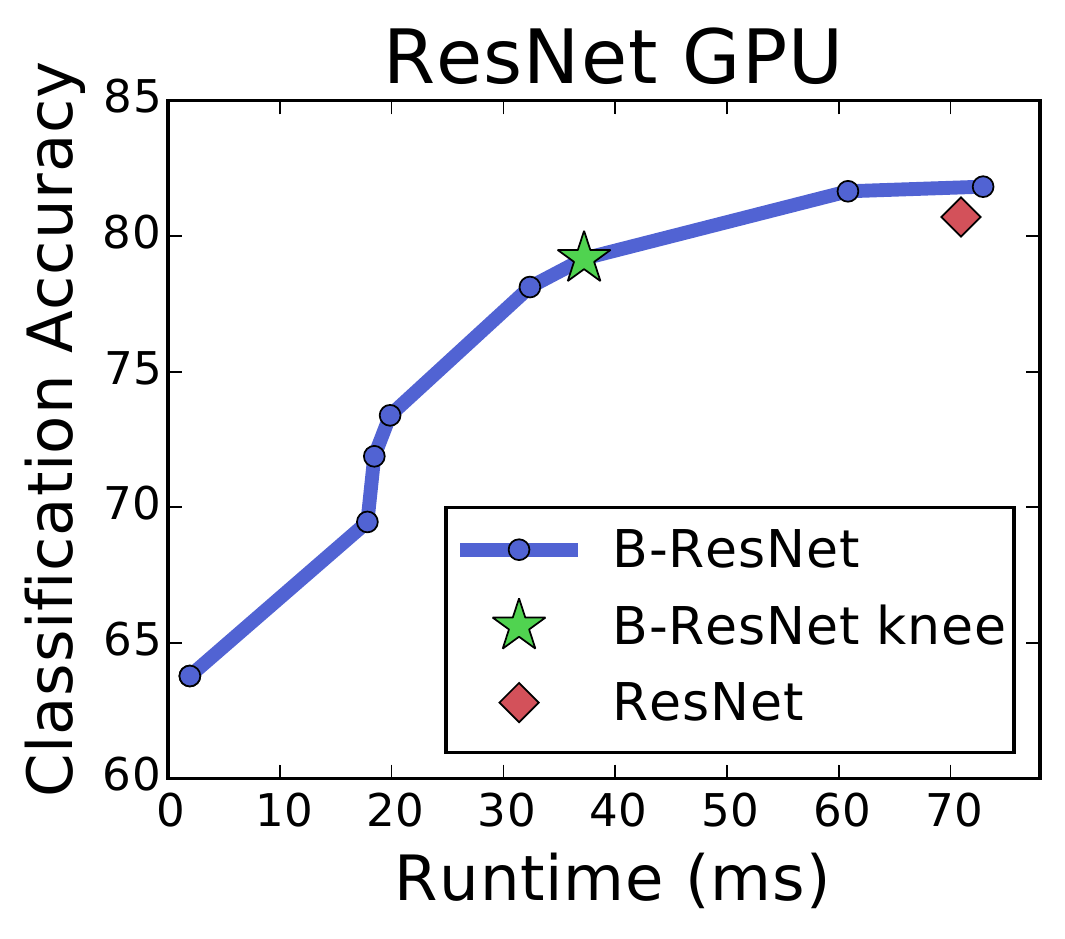}

\caption{GPU performance results for BranchyNet when applied to LeNet, AlexNet and ResNet. The original network accuracy and runtime are shown as the red diamond. The BranchyNet modification to each network is shown in blue. Each point denotes
different combinations of entropy thresholds for the branch exit points (found via sweeping over $\boldT$). The star denotes a knee point in the curve, with additional analysis shown in Table~\ref{table:perf}. The CPU performance results have similar characteristics, and can also be found in Table~\ref{table:perf}. Runtime is measured in milliseconds (ms) of inference per sample. For this evaluation, we use batch size 1 as evaluated in~\cite{han2015deep, kim2015compression} in order to target real-time streaming applications. A larger batch size allows for parallelism, but lessens the benefit of early exit as all samples in a batch must exit before a new batch can be processed.}
\label{fig:perf}
\end{minipage}
\hfill
\begin{minipage}[t]{.25\linewidth}

\centering
\includegraphics[width=1\linewidth]{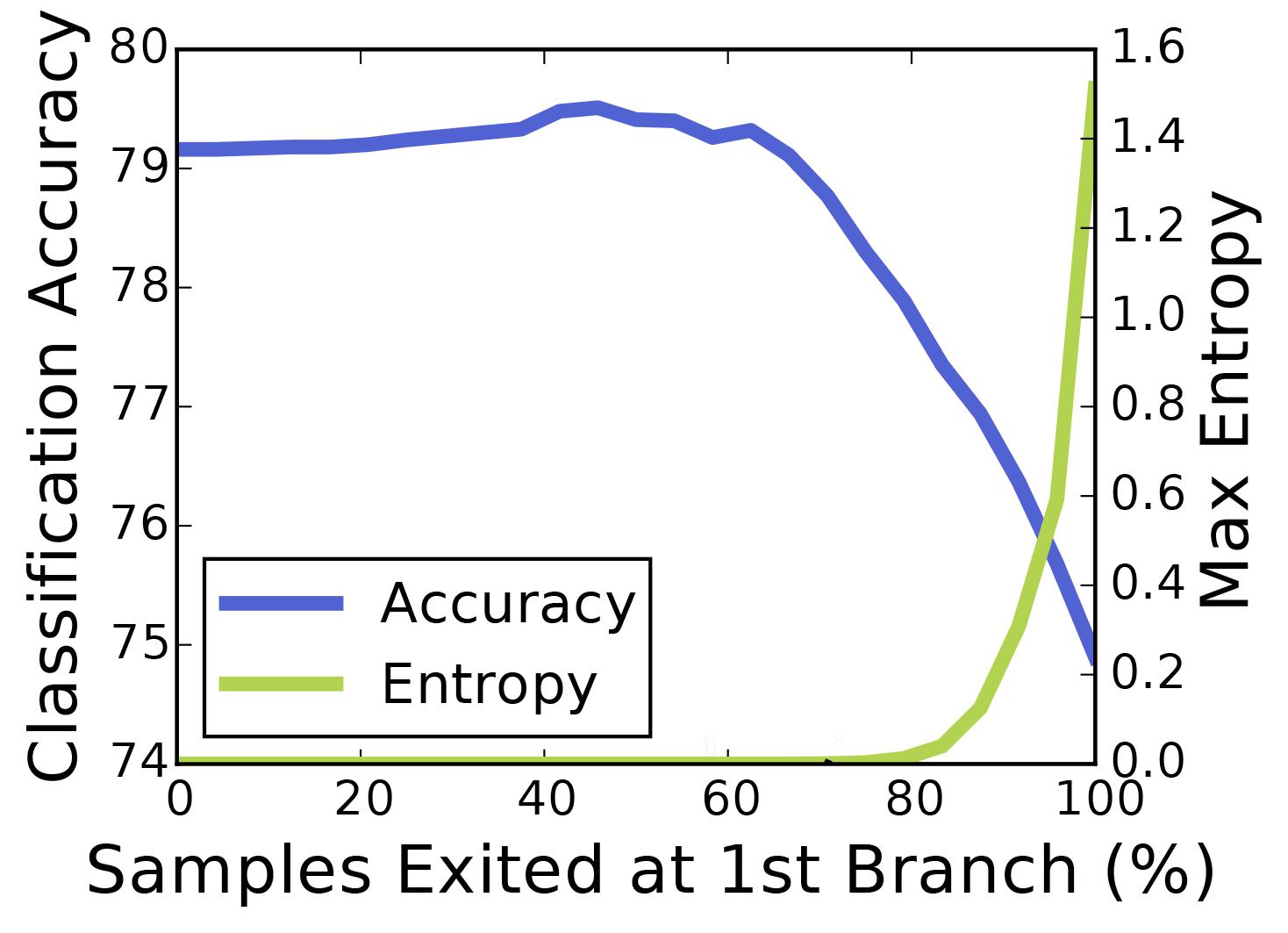}

\caption{The overall classification accuracy of B-AlexNet for varying entropy threshold for the first exit branch. For this experiment, all samples not exited in at the first branch are exited at the final exit. The entropy at a given value is the max entropy of all samples up to that point.}
\label{fig:entropy}

\end{minipage}

\end{figure*}

Figure~\ref{fig:perf} shows the GPU performance results of BranchyNet when applied to each network. For all of the networks, BranchyNet outperforms the original baseline network. The reported runtime is the average among all test samples. B-LeNet has the largest performance gain due to a more efficient branch which achieves almost the same level of accuracy as the last exit branch. For AlexNet and ResNet, we see that the performance gain is still substantial, but since more samples are required to exit at the last layer, smaller than B-LeNet. The knee point denoted as the green star represents an optimal threshold point, where the accuracy of BranchyNet is comparable to the main network, but the inference is performed significantly faster. For B-ResNet, the accuracy is slightly lower than the baseline. A different threshold could be chosen which gives accuracy higher than ResNet but with much less savings in inference time. The performance characteristics of BranchyNet running on CPU follow a similar trend to the performance of BranchyNet running on GPU.

Table~\ref{table:perf} highlights the selected knee threshold values, exit (\%) and gain in speed up, for BranchyNet for each network for both CPU and GPU. The $\boldT$ column denotes the threshold values for each exit branch. Since the last exit branch must exit all samples, it does not require an exit threshold. Therefore, for a 2-branch network, such as B-LeNet, there is a single $\boldT$ value and for a 3-branch network, such as B-AlexNet and B-ResNet, there are two $\boldT$ values. Further analysis of the sensitivity of the $\boldT$ parameters is discussed in Section~\ref{sec:discussion}. The Exit (\%) column shows the percentage of samples exited at each branch point. For all networks, we see that BranchyNet is able to exit a large percentage of the test samples before the last layer, leading to speedups in inference time. B-LeNet exits 94\% of samples at the first exit branch, while B-AlexNet and B-ResNet exit 65\% and 41\% respectively. Exiting these samples early translate to CPU/GPU speedup gains of 5.4/4.7x over LeNet, 1.5/2.4x over AlexNet, and 1.9/1.9x over ResNet.  The branch structure for B-ResNet mimics that of B-AlexNet.

\newcommand{\mulrow}[1]{\multirow{6}{*}{#1}}
\newcommand{\mulcol}[1]{\multicolumn{1}{l}{#1}}

\begin{table}[htp]
\centering
\caption{Selected performance results for BranchyNet on the different network structures. The BrachyNet rows correspond to the knee points denoted as green stars in Figure~\ref{fig:perf}.}
\label{table:perf}
\tabcolsep=0.11cm
\begin{tabular}{lllllll}
\hline
& \textbf{Network}        & \textbf{Acc.} (\%) & \textbf{Time} (ms) & \textbf{Gain} & \textbf{Thrshld. $\boldT$}     & \textbf{Exit} (\%)          \\ \hline
% \rowcolor{maroon!10}
\multirow{6}{*}{\textbf{CPU}}   & LeNet            &  99.20     & 3.37         & -       & -              & -              \\ 
% \rowcolor{maroon!10}
                                & B-LeNet          &  99.25    & 0.62         & 5.4x    & 0.025          & 94.3, 5.63       \\ 
                                & AlexNet        &  78.38    & 9.56         & -       & -              & -                  \\ 
                         & B-AlexNet &  79.19    & 6.32         & 1.5x    & 0.0001, 0.05 & 65.6, 25.2, 9.2  \\ 
% \rowcolor{maroon!10}
                         & ResNet         &  80.70    & 137.20       & -       & -              & -                  \\ 
% \rowcolor{maroon!10}
                         & B-ResNet  &  79.17    & 73.5         & 1.9x    & 0.3, 0.2     & 41.5, 13.8, 44.7 \\ \hline
                         & LeNet          &  99.20     & 1.58         & -       & -              & -                  \\ 
                         & B-LeNet      & 99.25    & 0.34         & 4.7x    & 0.025          & 94.3, 5.63      \\ 
% \rowcolor{maroon!10}
                         & AlexNet        &  78.38    & 3.15         & -       & -              & -                  \\ 
% \rowcolor{maroon!10}
                         & B-AlexNet &  79.19    & 1.30         & 2.4x    & 0.0001, 0.05 & 65.6, 25.2, 9.2  \\ 
                         & ResNet         &  80.70    & 70.9         & -       & -              & -                  \\ 
\multirow{-6}{*}{\textbf{GPU}}    & B-ResNet  &  79.17    & 37.2         & 1.9x    & 0.3, 0.2     & 41.5, 13.8, 44.7 \\ \hline
\end{tabular}
\end{table}

\section{Analysis and Discussion}
\label{sec:discussion}
In this section, we provide additional analysis on key aspects BranchyNet.

\subsection{Hyperparameter Sensitivity}
\label{sec:hyper}

Two important hyperparameters of BranchyNet are the weights ${w_n}$ in joint optimization (Section~\ref{sec:training}) and the exit thresholds $\boldT$ for the fast inference algorithm described in Figure~\ref{alg:branchy}. When selecting the weight of each branch, we observed that giving more weight to early branches improves the accuracy of the later branches due to the added regularization. On a simplified version of BranchyAlexNet with only the first and last branch, weighting the first branch with $1.0$ and the last branch with $0.3$ provides a 1\% increase in classification accuracy over weighting each branch equally. Giving more weight to earlier exit branches encourages more discriminative feature learning in early layers of the network and allows more samples to exit early with high confidence.

Figure~\ref{fig:entropy} shows how the choice of $\boldT$ affects the number of samples exited at the first branch point in B-AlexNet. We observe that the entropy value has a distinctive knee where it rapidly becomes less confident in the test samples. Thus in this case it is relatively easy to identify the knee and learn a corresponding threshold. In practice, the choice of exit threshold for each exit point depends on applications and datasets. The exit thresholds should be chosen such that it satisfies the inference latency requirement of an application while maintaining the required accuracy.

An additional hyperparameter not mentioned explicitly is the location of the branch points in the network. In practice, we find the location of the first branch point depends on the difficulty of the dataset. For a simpler dataset, such as MNIST, we can place a branch directly after the first layer and immediately see accurate classification. For more challenging datasets, branches should be placed higher in order to still achieve strong classification performance. For any additional branches, we currently place them at equidistant points throughout the network. Future work will be to derive an algorithm to find the optimal placement locations of the branches automatically.

\subsection{Tuning Entropy Thresholds}
\label{sec:thresholds}
The results shown in Figure~\ref{fig:perf} provides the accuracy and runtime for a range of $\boldT$ values. These $\boldT$ values show how BranchyNet trades off accuracy for faster runtime as the entropy thresholds increase. However, in practice, we may want to set $\boldT$ automatically to met a specified runtime or accuracy constraint. One approach is to simply screen over $\boldT$ as done here and pick a setting that satisfies the constraints. We provided code used to generate the performance results which includes a method for performing this screening~\cite{branchycode}.

Additionally, it may be possible to use a Meta-Recognition algorithm~\cite{scheirer2011meta, zhang2014predicting} to estimate the characteristics of unseen test samples and adjust $\boldT$ automatically in order to maintain a specified runtime or accuracy goal. One simple approach for creating such a Meta-Recognition algorithm would be to train a small Multilayer Perceptron (MLP) for each corresponding exit point on the output softmax probability vectors $\hat\boldy$ for that exit. The MLP at an exit point would attempt to predict if a given sample would be correctly classified at the specific exit. More generally, this approach is closely related to the open world recognition problem~\cite{bendale2015towards, bendale2015towards2}, which is interested in quantifying the uncertainty of a model for a particular set of unseen or out of set test samples. We can expand on the MLP approach further by using a different formulation than SoftMax, such as OpenMax~\cite{bendale2015towards}, which attempts to quantify the uncertainty directly in the probability vector $\hat\boldy$ by adding an additional uncertain class. These approaches could be used to tune $\boldT$ automatically to a new test set by estimating the difficulty of the test data and adapting $\boldT$ accordingly to meet the runtime or accuracy constraints. This work is outside the scope of this paper, which only provides the groundwork BranchyNet architecture, but will be explored in future work.

% \begin{figure}[htp]
% \centering
% \includegraphics[width=.7\linewidth]{entropy}

% \caption{The overall classification accuracy of B-AlexNet for different entropy threshold for the first exit branch. For this experiment, all samples not exited in at the first branch are exited at the final exit. The entropy at a given value is the max entropy of all samples up to that point.}
% \label{fig:entropy}
% \end{figure}

\subsection{Effects of Structure of Branches}
\label{sec:branches}

Figure~\ref{fig:num_layers} shows the impact on the accuracy of the last exit by adding additional convolutional layers in an earlier side branch for a modified version of B-AlexNet with only the first side branch. We see that there is a optimal number of layers to improve the accuracy of the main branch, and that adding too many layers can actually harm overall accuracy. In addition to convolutional layers, adding a few fully-connected layers after convolutional layers to a branch also proves helpful since this allows local and global features to combine and form more discriminative features. The number of layers in a branch and the size of an exit branch should be chosen such that the overall size of the branch is less than amount of computation needed to do to exit at a later exit point. Generally, we find that earlier branch points should have more layers, and later branch points should have fewer layers. 

\subsection{Effects of cache}
\label{sec:cache}
\begin{figure}[!htb]
\centering
    \begin{minipage}[t]{.49\linewidth}
        \centering
        \includegraphics[width=\linewidth]{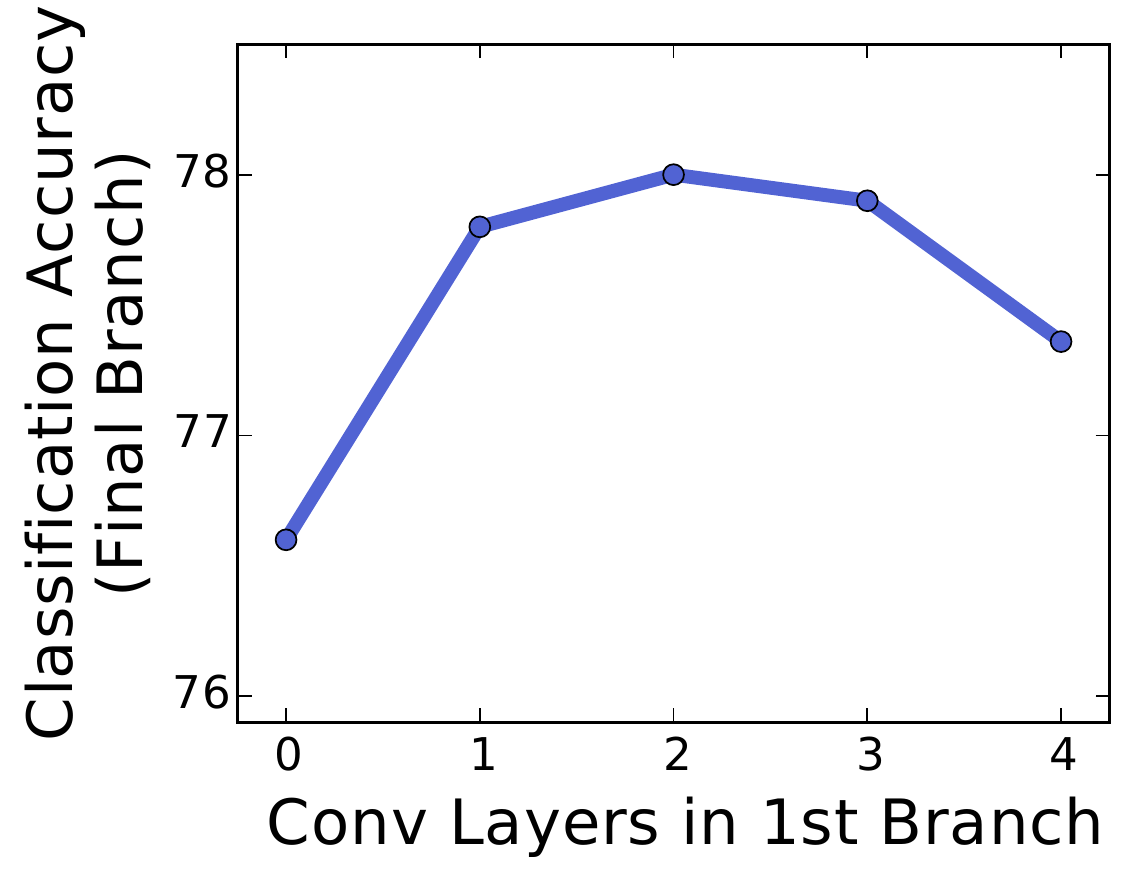}
        \caption{The impact of the number of convolutional layers in the first branch on the final branch classification accuracy for B-AlexNet.}
        \label{fig:num_layers}
    \end{minipage}
    \hfill
    \begin{minipage}[t]{.49\linewidth}
        \centering
        \includegraphics[width=\linewidth]{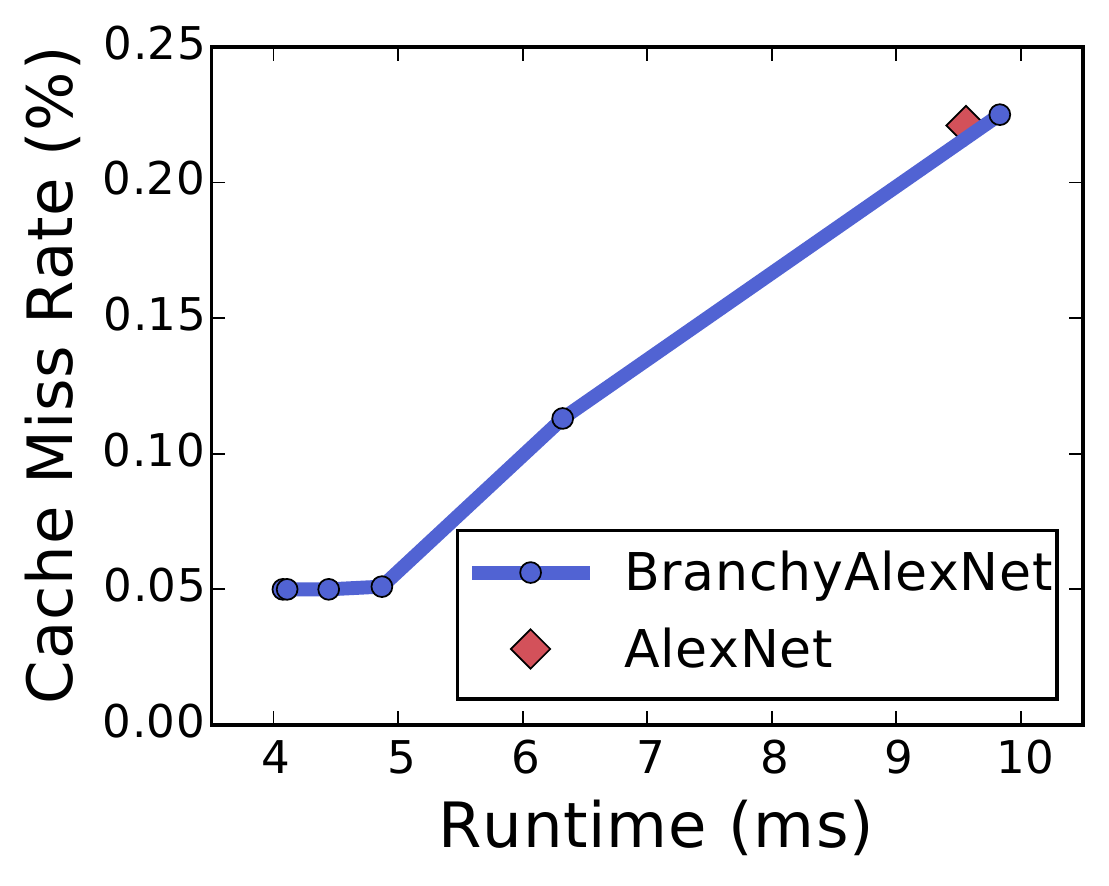}
        \caption{The runtime and CPU cache miss rate for the B-AlexNet model as the entropy threshold is varied.}
        \label{fig:alex_cache}
    \end{minipage}
\end{figure}

Since the majority of samples are exited at early branch points, the later branches are used more rarely. This allows weights at these early exit branches to be cached more efficiently. Figure~\ref{fig:alex_cache} shows the effect of cache based on various $\boldT$ values for B-AlexNet. We see that the more aggressive $\boldT$ values have faster runtime on the CPU and also less cache miss rates. One could use this insight to select a branch structure that can fits more effectively in a cache, potentially speeding up inference further.

\section{Conclusion}

We have proposed BranchyNet, a novel network architecture that promotes faster inference via early exits from branches. Through proper branching structures and exit criteria as well as joint optimization of loss functions for all exit points, the architecture is able to leverage the insight that many test samples can be correctly classified early and therefore do not need the later network layers. We have evaluated this approach on several popular network architectures and shown that BranchyNet can reduce the inference cost of deep neural networks and provide 2x-6x speed up on both CPU and GPU.

BranchyNet is a toolbox for researchers to use on any deep network models for fast inference. BranchyNet can be used in conjunction with prior works such as network pruning and network compression~\cite{buciluǎ2006model,han2015deep}. BranchyNet can be adapted to solve other types of problems such as image segmentation, and is not just limited to classification problems. For future work, we plan to explore Meta-Recognition algorithms, such as OpenMax, to automatically adapt $\boldT$ to new test samples.

%The only additional change is the exiting criteria which can be designed given the objective function of the inference.

% Future Works 

% Does the number of classes affect the exit profile of exit points in BranchyNet? Can exit thresholds that will maximize the overall accuracy be discovered automatically?

%

\section*{Acknowledgment}
  This work is supported in part by gifts from the Intel Corporation
  and in part by the Naval Supply Systems Command award under the
  Naval Postgraduate School Agreements No. N00244-15-0050
  and No. N00244-16-1-0018.
% The authors would like to thank...

\footnotesize
\bibliographystyle{IEEEtran}
\bibliography{IEEEabrv,sigproc}

\end{document}